\documentclass[11pt,a4paper]{article}
\usepackage[hyperref]{acl2019}
\usepackage{times}
\usepackage{latexsym}
\usepackage{url}
\usepackage{graphicx}
\usepackage{times}
\usepackage{booktabs}
\usepackage{todonotes}
\usepackage{multirow}
\usepackage{microtype}

\usepackage{newtxmath}

\DeclareMathAlphabet{\mathcal}{OMS}{cmsy}{m}{n}
\DeclareMathAlphabet{\mathbb}{U}{msb}{m}{n}
\usepackage{dingbat}

\usepackage{tabularx}
\usepackage{float}
\usepackage{adjustbox}
\usepackage{makecell}
\usepackage{caption}
\usepackage{subfigure}
\aclfinalcopy % Uncomment this line for the final submission
\def\aclpaperid{1729} %  Enter the acl Paper ID here

\newcommand{\affhit}{\textsuperscript{\dag}}
\newcommand{\affjhu}{\textsuperscript{\ddag}}

\newcommand{\bn}[1]{\textbf{(#1)}}

\title{Learning to Rank for Plausible Plausibility}

\author{
  Zhongyang Li\affhit\affjhu\thanks{~ This work was done while the first author was visiting Johns Hopkins University.} \quad  Tongfei Chen\affjhu \quad  Benjamin Van Durme\affjhu \\
  \affhit ~ Harbin Institute of Technology \\
  \affjhu ~ Johns Hopkins University \\
  { \tt zyli@ir.hit.edu.cn, \string{tongfei,vandurme\string}@cs.jhu.edu} \\
}

\date{}

\begin{document}
\maketitle

\begin{abstract}
Researchers illustrate improvements in contextual encoding strategies via resultant performance on a battery of shared Natural Language Understanding (NLU) tasks. Many of these tasks are of a categorical prediction variety: given a conditioning context (e.g., an NLI \emph{premise}), provide a label based on an associated prompt (e.g., an NLI \emph{hypothesis}). The categorical nature of these tasks has led to common use of a cross entropy log-loss objective during training. We suggest this loss is intuitively wrong when applied to \emph{plausibility} tasks, where the prompt by design is neither categorically \emph{entailed} nor \emph{contradictory} given the context. Log-loss naturally drives models to assign scores near 0.0 or 1.0, in contrast to our proposed use of a margin-based loss. Following a discussion of our intuition, we describe a confirmation study based on an extreme, synthetically curated task derived from MultiNLI. We find that a margin-based loss leads to a more plausible model of plausibility. Finally, we illustrate improvements on the Choice Of Plausible Alternative (COPA) task through this change in loss.

\end{abstract}

\section{Introduction}

Contextualized encoders such as GPT \cite{radford2018improving} and BERT \cite{devlin2018bert} have led to improvements on various structurally similar Natural Language Understanding (NLU) tasks such as variants of Natural Language Inference (NLI).  Such tasks model the conditional interpretation of a sentence (e.g., an NLI \emph{hypothesis}) based on some other context (usually some other sentence, e.g., an NLI \emph{premise}).  The structural similarity of these tasks points to a structurally similar modeling approach: (1) concatenate the conditioning context (premise) to a sentence to be interpreted, (2) \emph{read} this pair using a contextualized encoder, then (3) employ the resultant representation to support classification under the label set of the task.  NLI datasets employ a categorical label scheme (\emph{Entailment}, \emph{Neutral}, \emph{Contradiction}) which has led to the use of a cross-entropy log-loss objective at training time: learn to maximize the probability of the correct label, and thereby  minimize the probability of the competing labels.

\begin{figure}
    \centering
   \begin{tabular}{|ll|l|}
\hline
$p$ & \emph{I just stopped where I was} & \\
\hline\hline
$h_{\rm E}$ & \emph{I stopped in my tracks} & \checkmark \\
$h_{\rm N}$ &  \emph{I stopped running right were I was} &  \\
\hline
$h_{\rm N}$ & \emph{I stopped running right were I was} & \checkmark \\
$h_{\rm C}$ & \emph{I continued on my way} & \\
\hline
\end{tabular}
    \caption{COPA-like pairs may be constructed from datasets such as MultiNLI, where a \emph{premise} and two \emph{hypotheses} are presented, where the correct -- most plausible --  item depends on the competing hypothesis.}
    \label{fig:copa-mnli}
\end{figure}

\begin{figure}
    \centering
    \includegraphics[width=0.95\columnwidth]{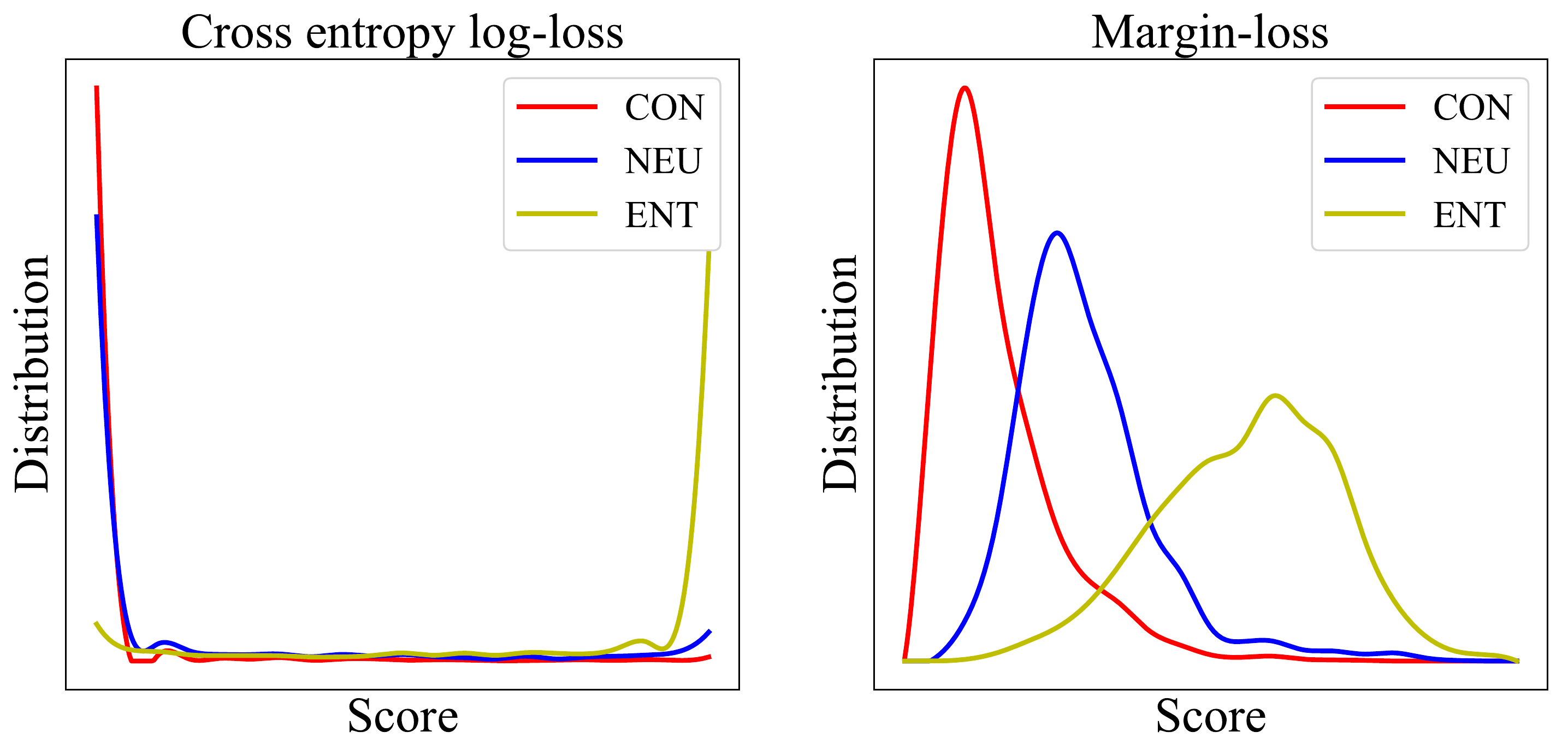}
    % \vspace{-0.3cm}
    \caption{Dev set score distribution on COPA-pairs derived from MNLI, after training with cross entropy log-loss and margin-loss.  Margin-loss leads to a more intuitively \emph{plausible} encoding of \emph{Neutral} statements.}
    \label{fig:task}
\end{figure}

We suggest that this approach is intuitively problematic when applied to a task such as COPA (Choice Of Plausible Alternative) by \newcite{roemmele2011choice}, where one is provided with a premise and two or more alternatives, and the model must select the most sensible hypothesis, with respect to the premise \emph{and the other options}. As compared to NLI datasets, COPA was designed to have alternatives that are neither strictly true nor false in context: a procedure that maximizes the probability of the correct item at training time, thereby minimizing the probability of the other alternative(s), will seemingly learn to \emph{misread} future examples.

We argue that COPA-style tasks should intuitively be approached as \emph{learning to rank} problems~\cite{burges2005learning,cao2007learning}, where an encoder on competing items is trained to assign \emph{relatively} higher or lower scores to candidates, rather than maximizing or minimizing probabilities.  In the following we investigate three datasets, beginning with a constructed COPA-style variant of MultiNLI \cite[later MNLI]{mnli}, designed to be adversarial (see~\autoref{fig:copa-mnli}). 
%For example, at training time a model is exposed to a particular premise and then required to predict whether the \emph{Entailed} or \emph{Neutral} hypothesis is more plausible.  This would seemingly be an easy problem under contemporary models, but the model is later required to perform this task when presented the identical premise, the identical \emph{Neutral} hypothesis, and then the previously unseen \emph{Contradiction} from MNLI.  This proof of concept  demonstrates that traditional cross entropy based models suffer from pushing the scores to be 0 or 1, whereas our proposed margin-based methods can learn to assign more plausible plausibility scores for  hypotheses, as is shown in \autoref{fig:task}.
%
Results on this dataset support our intuition (see~\autoref{fig:task}).  We then construct a second synthetic dataset based on JOCI \cite{zhang2017ordinal}, which employed a finer label set than NLI, and a margin-based approach strictly outperforms log-loss in this case.  Finally, we demonstrate state-of-the-art on COPA, showing that a BERT-based model trained with margin-loss significantly outperforms a log-loss alternative.

\section{Background}

%Commonsense causal reasoning problems are sometimes formulated as a \emph{plausibility} task -- given a premise or context, of all the candidate hypotheses, which one is the most plausible? This line of research led to the COPA evaluation \cite{roemmele2011choice}, where a model is expected to choose the more plausible hypothesis from 2 choices. A series of work has been done addressing the plausibility task: by causality estimation through pointwise mutual information \cite{gordon2011commonsense} or data-driven methods \cite{luo2016commonsense, sasaki2017handling}, or through a pre-trained language model \cite[GPT]{radford2018improving}\footnote{~As is reported in \url{https://blog.openai.com/language-unsupervised/}.} that achieved state-of-the-art results.

A series of efforts have considered COPA: by causality estimation through pointwise mutual information \cite{gordon2011commonsense} or data-driven methods \cite{luo2016commonsense, sasaki2017handling}, or through a pre-trained language model \cite[GPT]{radford2018improving}.\footnote{~As reported in \url{https://blog.openai.com/language-unsupervised/}.}

Under the Johns Hopkins Ordinal Common-sense Inference (JOCI) dataset~\cite{zhang2017ordinal}, instead of selecting which hypothesis is the most plausible, a model is expected to directly assign ordinal 5-level Likert scale judgments (from \emph{impossible} to \emph{very likely}).  If taking an ordinal interpretation of NLI, this can be viewed as a 5-way variant of the 3-way labels used in SNLI~\cite{bowman2015large} and MNLI~\cite{mnli}. % Recent advancements in pre-trained language models \cite[BERT]{devlin2018bert} achieved state-of-the-art results on MultiNLI (MNLI) \cite{mnli}.

%The task of NLI is in a similar vein: given a premise and a hypothesis, a model should produce their logical relation: whether it be entailment, neutral or contradiction \cite{bowman2015large,conneau2017supervised}. Recent advancements in pre-trained language models \cite[BERT]{devlin2018bert} achieved state-of-the-art results on MultiNLI (MNLI) \cite{mnli}.

In this paper, we recast MNLI and JOCI as COPA-style plausibility tasks by sampling and constructing $(p, h, h^\prime)$ triples from these two datasets. Each premise-hypothesis pair $(p,h)$ is labeled with different levels of plausibility $y_{p,h}$.\footnote{~For MNLI, \emph{entailment} $>$ \emph{neutral} $>$ \emph{contradiction}; for JOCI, \emph{very likely} $>$ \emph{likely} $>$ \emph{plausible} $>$ \emph{technically possible} $>$ \emph{impossible}.} 

%\noindent \textbf{Contextualized encoders}
\section{Models}
In models based on GPT and BERT for plausibility or NLI, similar neural architectures have been employed. The premise $p$ and hypothesis $h$ are concatenated into a sequence with a special delimiter token, along with a special sentinel token \textsc{cls} inserted as the token for feature extraction:
 \begin{align*}
   \mathrm{BERT: \quad} &\left[ \textsc{cls} ~;~ p ~;~ \textsc{sep} ~;~ h ~;~ \textsc{sep} \right] \\
   \mathrm{GPT: \quad} &\left[ \textsc{bos} ~;~ p ~;~ \textsc{eos} ~;~ h ~;~ \textsc{cls} \right]
 \end{align*}

The concatenated string is passed into the BERT or GPT encoder. One takes the encoded vector of the \textsc{cls} state as the feature vector extracted from the $(p, h)$ pair. Given the feature vector, a dense layer is stacked upon to get the final score $F(p, h)$, where $F: \mathcal{P} \times \mathcal{H} \to \mathbb{R}$ is the model. 

\paragraph{Cross entropy loss} The model is trained to maximize the probability of the correct candidate, normalized over all candidates in the set (leading to a cross entropy log-loss between the posterior distribution of the scores and the true labels):
\begin{equation} 
    P(h_i \mid p) = \dfrac{\exp F(p, h_i)}{\displaystyle\sum_{j=1}^{N} \exp F(p, h_j)} \ .
\label{eq:log}
\end{equation}
\vspace{-0.3cm}

\paragraph{Margin-based loss}  As we have argued before, the cross entropy loss employed in \autoref{eq:log} is problematic. 
% \todo[inline]{argue about ranking instead of classification} 
Instead we propose to use the following margin-based triplet loss \cite{DBLP:conf/esann/WestonW99,chechik2010large,li2018constructing}:
      \begin{equation} 
        L = \frac{1}{N} \sum_{h > h^\prime} \max \{0, \xi - F(p, h) + F(p, h^\prime) \} \ ,
      \end{equation}
where $N$ is the number of pairs of hypotheses where the first is more plausible than the second under the given premise $p$; $h > h^\prime$ means that $h$ ranks higher than (i.e., is more plausible than) $h^\prime$ under premise $p$; and $\xi$ is a margin hyperparameter denoting the desired scores difference between these two hypotheses.

\section{Recasting Datasets}

We consider three datasets: MNLI, JOCI, and COPA. % \citep{roemmele2011choice}, MNLI \citep{mnli}, and JOCI \citep{zhang2017ordinal}.
These are all cast as plausibility datasets, into a format comprising $(p, h, h^\prime)$ triples, where $h$ is more plausible than $h^\prime$ under the context of premise $p$.% as we have discussed.

\paragraph{MNLI} In MNLI, each premise $p$ is paired with 3 hypotheses. We cast the label on each hypothesis as a relative plausibility judgment, where \emph{entailment} $>$ \emph{neutral} $>$ \emph{contradiction} (we label them as 2, 1, and 0). We construct two 2-choice plausibility tasks from MNLI: 
\begin{align*}
\mathrm{MNLI}_1 &= \{ (p, h, h^\prime) \mid y_{p, h} > y_{p, h^\prime} \} \\
\mathrm{MNLI}_2 &= \{ (p, h, h^\prime) \mid (y_{p, h}, y_{p, h^\prime}) \in \{ (2,1), (1,0)\} \}
\end{align*}

$\mathrm{MNLI}_1$ comprises all pairs labeled with 2/1, 2/0, or 1/0; whereas $\mathrm{MNLI}_2$ removes the presumably easier 2/0 pairs. For $\mathrm{MNLI}_1$, the training set is constructed from the original MNLI training dataset, and the dev set for $\mathrm{MNLI}_1$ is derived from the original MNLI matched dev dataset. For $\mathrm{MNLI}_2$, all of the examples in our training and dev sets is taken from the original MNLI training dataset, hence the same premise exists in both training and dev. This is by our adversarial design: each neutral hypothesis appears either as the preferred (beating contradiction), or dispreferred alternative (beaten by entailment), which is flipped at evaluation time.

% Actually, $\mathrm{MNLI}_2$ is considered as an adversarial dataset constructed in such a way.
% Since the hypotheses in MNLI is labeled as three polar categories, the two recast MNLI datasets can be seen as synthetic plausibility datasets. 

\paragraph{JOCI} In JOCI, every inference pair is labeled with their ordinal inference Likert-scale labels 5, 4, 3, 2, or 1. Similar to MNLI, we cast these to 2-choice problems under the following conditions:
\begin{align*}
  \mathrm{JOCI}_1 &= \{ (p, h, h^\prime) \mid y_{p, h} > y_{p, h^\prime} \ge 3 \} \\
  \mathrm{JOCI}_2 &= \{ (p, h, h^\prime) \mid (y_{p, h}, y_{p, h^\prime}) \in \{ (5,4), (4,3)\} \}
\end{align*}
% \vspace{-0.5cm}

We ignore inference pairs with scores below 3, aiming for sets akin to COPA, where even the dis-preferred option is still often semi-plausible. %The training and development examples for these two JOCI datasets are constructed similar to the two MNLI datasets. 
% Each hypothesis we take is plausible but with different plausibility. Hence, the two JOCI datasets can be seen as synthetic real plausibility datasets. 

\paragraph{COPA} We label alternatives as 1 (the more plausible one) and 0 (otherwise). The original dev set in COPA is used as the training set.
% COPA is used as the real evaluation dataset to demonstrate the advantage of the margin-loss.

\autoref{tab:datasets} shows the statistics of these datasets.

\begin{table}[t]
    \centering
    \adjustbox{max width=\textwidth}{
    \begin{tabular}{lcc}
         \toprule
         Dataset        & Train & Eval \\  
         \midrule
         MNLI$_1$       &  410k & {\bf dev:} 8.2k \\
         MNLI$_2$       &  142k  & {\bf dev:} 130k  \\ 
         JOCI$_1$       &  8.7k & {\bf dev:} 3.0k \\
         JOCI$_2$       &  2.3k & {\bf dev:} 1.9k  \\ 
         COPA           &  500  & {\bf test:} 500 \\
         \bottomrule 
    \end{tabular}
    }
    \caption{Statistics of various plausibility datasets. All numbers are numbers of $(p, h, h^\prime)$ triplets. }
    \label{tab:datasets}
    \vspace{-0.3cm}
\end{table}

\section{Experiments and Analyses}

\noindent \textbf{Setup}
We fine-tune the BERT-\textsc{base-uncased} \citep{devlin2018bert} using our proposed margin-based loss, and perform hyperparameter search on the margin parameter $\xi$. 

For the recast MNLI and JOCI datasets, the margin hyperparameter $\xi=0.2$. Since COPA does not have a training set, we use the original dev set as the training set, and perform 10-fold cross validation to find the best hyperparameter $\xi=0.37$. We employ the Adam optimizer \cite{kingma2014adam} with initial learning rate $\eta=3\times10^{-5}$, fine-tune for at most 3 epochs and use early-stopping to select the best model.

\paragraph{Results on Recast MNLI and JOCI}

Table \ref{tab:mnli_joci_results} shows results on the recast MNLI and JOCI datasets. We find that for the two synthetic MNLI datasets, margin-loss performs similarly to cross entropy log-loss.  Shifting to the JOCI datasets, with less extreme (contradiction / entailed) hypotheses, especially in the adversarial JOCI$_{2}$ variant, margin-loss outperforms log-loss.

\begin{figure*}[t!]
  \centering 
  \subfigure[MNLI$_1$]{ 
    \label{fig:subfig:a}
    \includegraphics[width=0.99\columnwidth]{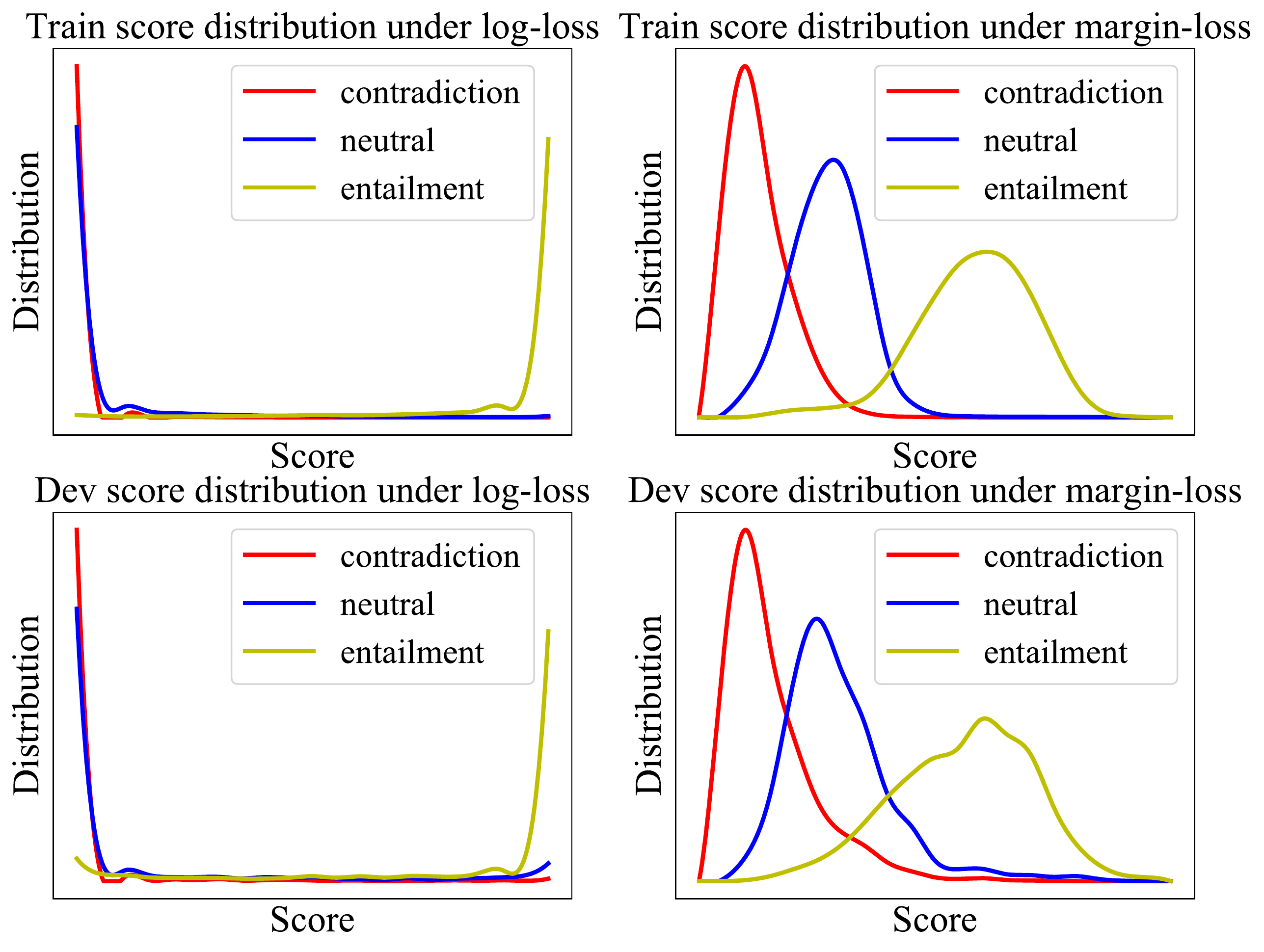}
  }
  \hspace{1ex}
  \subfigure[JOCI$_1$]{ 
    \label{fig:subfig:b}
    \includegraphics[width=0.99\columnwidth]{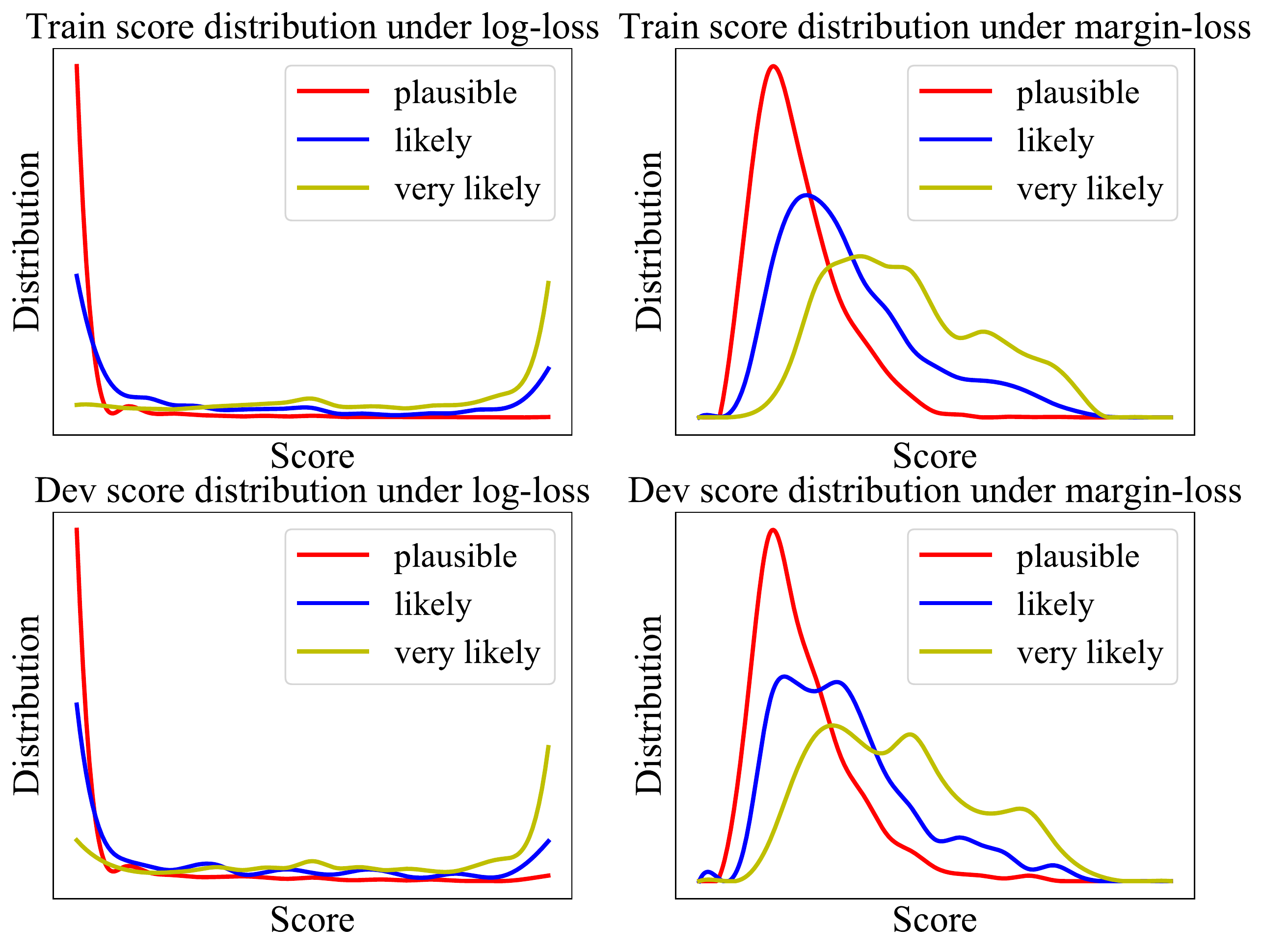}
  }
  \vspace{-0.3cm}
  \caption{Train and dev score distribution after training with a cross entropy log-loss and a margin-loss.} 
  \label{fig:subfig}
\end{figure*}

Though log-loss and margin-loss give close quantitative results on predicting the more plausible $(p,h)$ pairs, they do so in different ways, confirming our intuition. From Figure \ref{fig:subfig} we find that the log-loss always predicts the more plausible $(p,h)$ pair with very high probabilities close to 1, and predicts the less plausible $(p,h)$ pair with very low probabilities close to 0. %This may be acceptable in the synthetic MNLI datasets, but it is unreasonable and very weird in the real plausibility tasks, where all the hypotheses are likely and with different probabilities higher than 0. 
Figure \ref{fig:subfig}, showing a per-premise normalized score distribution from margin-loss, is more reasonable and explainable: hypotheses with different plausibility are distributed hierarchically between 0 and 1.

\begin{table}[t]
    \centering
    \begin{tabular}{lcc} 
         \toprule
         Dataset        & Log loss & Margin loss \\ 
         \midrule
         MNLI$_1$       &   \textbf{93.6}  & 93.4 \\
         MNLI$_2$       &   87.9  &   87.9    \\ 
         JOCI$_1$       &   86.6  &  \textbf{86.9}    \\
         JOCI$_2$       &   76.6  &  \textbf{78.0} \\ 
         \bottomrule 
    \end{tabular}
    \caption{Results on recast MNLI and JOCI.}
    \label{tab:mnli_joci_results}
    \vspace{-0.3cm}
\end{table}

\begin{table}[H] 
    \centering
    \begin{tabular}{lc} 
         \toprule
         Method        &  Acc (\%) \\  
         \midrule
         PMI \cite{jabeen2014using} & 58.8 \\
         PMI\_EX \cite{gordon2011commonsense} & 65.4 \\
         CS \cite{luo2016commonsense}& 70.2 \\
         CS\_MWP \cite{sasaki2017handling} & 71.2 \\
         \midrule
         BERT$_{\text{log}}$ (ours) &  73.4    \\
         BERT$_{\text{margin}}$ (ours)& \textbf{75.4}  \\
         \bottomrule 
    \end{tabular}
    \caption{Experimental results on COPA test set.}
    \label{tab:copa_results}
    \vspace{-0.3cm}
\end{table}

\begin{table*}[t!] \small
  \centering
  \begin{adjustbox}{max width=\textwidth}
  \begin{tabular}{cllccc}
    \toprule
    \bf Dataset & \bf Premise & \bf Hypotheses & \bf Gold & \bf Log & \bf Margin \\
    \midrule 
    \multirow{3}{*}{MNLI$_1$} & \multirow{3}{*}{\bn{1} \it I just stopped where I was.} & \bn{a} \it I stopped in my tracks & 2 & 0.919 & 0.568 \\
                          & & \bn{b} \it I stopped running right where I was. & 1 & 0.0807 & 0.358 \\
                          & & \bn{c} \it I continued on my way. & 0 & 1.71$\times 10^{-8}$ & 0.0739 \\
    \midrule
    \multirow{5}{*}{MNLI$_1$} & \multirow{5}{*}{\makecell[l]{\bn{2} \it An organization's activities, core \\ \it processes and resources must be \\ \it aligned to support its mission and \\ \it help it achieve its goals. }} &   \makecell[l]{ \bn{a} \it An organization is successful if its \\ \it activities, resources and goals align.} & 2 & 0.505 & 0.555 \\
                          & &  \makecell[l]{\bn{b} \it Achieving organizational goals \\ \it reflects a change in core processes.} & 1 & 0.495 & 0.257 \\
                          & &  \makecell[l]{\bn{c} \it A company's mission can be realized \\ \it even without the alignment of resources.} & 0 & 3.48$\times 10^{-5}$ & 0.187 \\
    \midrule
    \multirow{3}{*}{JOCI$_1$} & \multirow{3}{*}{\makecell[l]{\bn{3} \it A few people and cars out on \\ \it their daily commute on a rainy day.}} & \bn{a} \it The commute is a journey. & 5 & 0.994 & 0.473 \\
                          & & \bn{b} \it The commute is bad. & 4 & 5.79$\times 10^{-3}$ & 0.230 \\
                          & & \bn{c} \it The commute becomes difficult. & 3 & 1.28$\times 10^{-3}$ & 0.157 \\
    \midrule
    \multirow{5}{*}{JOCI$_1$} & \multirow{5}{*}{\makecell[l]{\bn{4} \it Cheerleaders in red uniforms \\ \it perform a lift stunt.}} & \bn{a}  \it The stunt is a feat. & 5 & 0.508 & 0.304 \\
                          & & \bn{b} \it The stunt is no fluke. & 4 & 0.486 & 0.279 \\
                          & & \bn{c} \it The stunt is dangerous. & 3 & 2.72$\times 10^{-4}$ & 0.166 \\
                          & & \bn{d} \it The stunt is remarkable. & 3 & 4.13$\times 10^{-3}$ & 0.153 \\
                          & & \bn{e} \it The stunt backfires. & 3 & 2.36$\times 10^{-4}$ & 0.107 \\
    \midrule
    \midrule
    \multirow{2}{*}{COPA} & \bn{5}  \it She jumped off the diving board. & \multirow{2}{*}{{\bn{a} \it The girl landed in the pool.}}  & 1 & 0.972 & 0.520\\
                          &  \bn{5$^{\boldsymbol{\prime}}$}  \it She ran on the pool deck. & & 0 & 0.028 & 0.480 \\
    \midrule
    \multirow{2}{*}{COPA} & \multirow{2}{*}{\makecell[l]{\bn{6} \it The student knew the answer \\ \it to the question.}} & \bn{a}  \it He raised his hand. & 1 & 0.982 & 0.738 \\
                          & & \bn{b}  \it He goofed off. & 0 & 0.018 &0.262 \\
    \bottomrule
  \end{tabular}
  \end{adjustbox}
  \caption{Examples of premises and their corresponding hypotheses in various plausibility datasets, with gold labels and scores given by the log-loss and margin-loss trained models.}
  \label{tab:examples}
\end{table*}

\paragraph{Results on COPA}

Table \ref{tab:copa_results} shows our results on COPA. Compared with previous state-of-the-art knowledge-driven baseline methods, a BERT model trained with a log-loss achieves better performance. When training the BERT model with a margin-loss instead of a  log-loss, our method gets the new state-of-the-art result on the established COPA splits, with an accuracy of 75.4\%.\footnote{~We exclude a blog-posted GPT result, which comes without experimental conditions and is not reproducible.}

\paragraph{Analyses}
\autoref{tab:examples} shows some examples from the MNLI$_1$, JOCI$_1$ and COPA datasets, with scores normalized with respect to all hypotheses given a specific premise. %(for sake of analysis across examples; relative magnitude of alternatives given a premise is unchanged). 

For the premise \bn{1} from MNLI$_1$, log-loss results in a very high score (0.919) for the entailment hypothesis \bn{1a}, while assigning a low score (0.0807) for the neutral hypothesis \bn{1b}, and an extremely low score (1.71$\times 10^{-8}$) for the contradiction hypothesis \bn{1c}. Though the log-loss can achieve high accuracy by making these extreme prediction scores, we argue these scores are unintuitive. For the premise \bn{2} from MNLI$_1$, log-loss again gives a very high score (0.505) for the hypothesis \bn{2a}. But it also gives a high score (0.495) for the neutral hypothesis \bn{2b}. The contradiction hypothesis \bn{2c} still gets an extremely low score (3.48$\times 10^{-5}$). 

These are the two ways for the log-loss approach to make predictions with high accuracy: always giving very high score for the entailment hypothesis and low score for the contradiction hypothesis, but giving either very high or very low score for the neutral hypothesis. In contrast, the margin-loss gives more intuitive scores for these two examples. Also, we get similar observations from the JOCI$_1$ examples \bn{3} and \bn{4}.

Example \bn{5} from COPA is asking for a more plausible \emph{cause} premise for the \emph{effect} hypothesis. Here, each of the two candidate premises \bn{5} and \bn{5$^{\boldsymbol{\prime}}$} is a possible answer. The log-loss gives very high (0.972) and very low (0.028) scores for the two candidate premises, which is unreasonable. Whereas the margin-loss gives much more rational ranking scores for them (0.52 and 0.48). For example \bn{6}, which is asking for a more likely effect hypothesis for the cause premise, margin-loss still gets more reasonable prediction scores than the log-loss.

Our qualitative analysis is related to the concept of \emph{calibration} in statistics: are these resulting scores close to their class membership probabilities? Our intuitive qualitative results might be thought as a type of calibration for the plausibility task (more ``reliable'' scores) instead of the more common multi-class classification \cite{zadrozny2002transforming,hastie1998classification,niculescu2005predicting}. 

\section{Conclusion}
In this paper, we propose that margin-loss in contrast to log-loss is a more plausible training objective for COPA-style \emph{plausibility} tasks.   Through adversarial construction we illustrated that a log-loss approach can be driven to encode plausible statements (Neutral hypotheses in NLI) as either extremely likely or unlikely, which was highlighted in contrasting figures of per-premise normalized hypothesis scores.
This intuition was shown to lead to a new state-of-the-art in the original COPA task, based on a margin-based loss.

%Though a margin-loss doesn't outperform log-loss on the synthetic MNLI dataset, it performs better on the real plausibility JOCI datasets. 
%Illustration of the score distribution suggests that the log-loss and margin-loss models predict from very different ways. 
%Finally, we achieve new state-of-the-art performance on the COPA task with a margin-loss.

% \newpage
\section*{Acknowledgements}
This work was partially sponsored by the China Scholarship Council.
It was also supported in part by DARPA AIDA. The authors thank the reviewers for their helpful comments.

\bibliography{acl2019}
\bibliographystyle{acl_natbib}
\end{document}